\documentclass[]{spie}  

\usepackage{amsmath,amsfonts,amssymb}
\usepackage{graphicx}
\usepackage[colorlinks=true, allcolors=blue]{hyperref}

\title{Generating Visual Stimuli from EEG Recordings using Transformer-encoder based EEG encoder and GAN}

\author{Rahul Mishra}

\author{Arnav Bhavsar}
\affil{Multimedia Analytics, Networks and Systems Lab\\School of Computing \& Electrical Engineering, IIT Mandi, Mandi, India}

\authorinfo{Further author information: (Send correspondence to Rahul Mishra)\\Rahul Mishra: E-mail: d16043@students.iitmandi.ac.in}

\begin{document} 
\maketitle

\begin{abstract}
In this study, we tackle a modern research challenge within the field of perceptual brain decoding, which revolves around synthesizing images from EEG signals using an adversarial deep learning framework. The specific objective is to recreate images belonging to various object categories by leveraging EEG recordings obtained while subjects view those images. To achieve this, we employ a Transformer-encoder based EEG encoder to produce EEG encodings, which serve as inputs to the generator component of the GAN network. Alongside the adversarial loss, we also incorporate perceptual loss to enhance the quality of the generated images.
\end{abstract}


\section{Introduction}
Brain decoding entails extracting insights regarding emotional states, abnormal brain functions, cognitive functions, or perceptual stimuli from typically non-invasive assessments of brain activity. Electroencephalography (EEG) serves as a prevalent and cost-effective method for measuring electrical brain activity~\cite{chen2014}. While historically prevalent in clinical research, the analysis of EEG signals has expanded in recent years, particularly with the rise in popularity of brain-computer interfaces for cognitive and perceptual tasks~\cite{mishra2021eeg}.

\begin{figure}[h!]
    \centering
    \includegraphics[scale = 0.4]{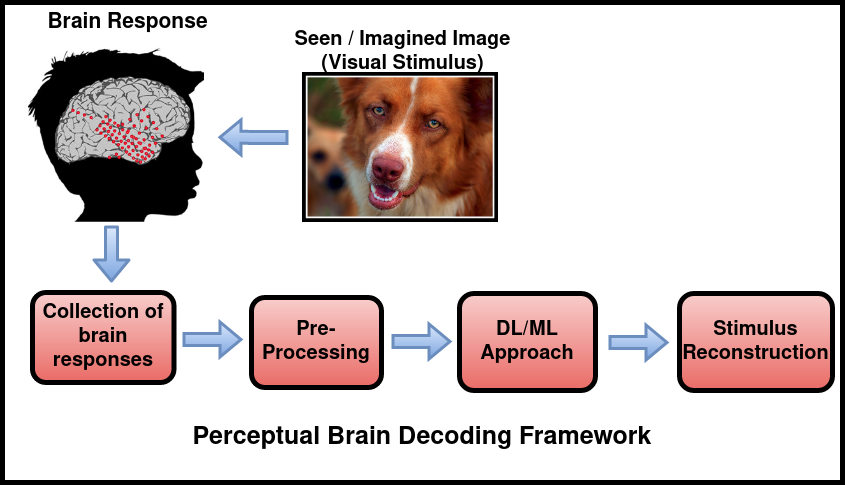}
    \caption{PBD Framework}
    \label{pbd}
\end{figure}

Perceptual Brain Decoding (PBD) is an approach depicted in Fig.~\ref{pbd}, leveraging brain responses evoked by diverse stimuli to discern the original perceptual stimulus, such as visual or auditory cues, or certain characteristics thereof. Generally, PBD offers advantages from both cognitive and clinical standpoints. Through PBD, diverse brain activity patterns corresponding to external stimuli can be scrutinized. In clinical settings, brain decoding techniques hold potential for communication with individuals experiencing conditions like locked-in syndrome or paralysis, which may impair motor and vocal functions. In such scenarios, attempts can be made to reconstruct the individual's responses or imaginations while providing perceptual stimuli. Additionally, improved methods in perceptual brain decoding could benefit applications such as memory retrieval or visualizing thoughts, aiding cognitive research and rehabilitation efforts.

Compared to various other EEG-based applications, such as identifying mental states corresponding to different emotions, intentions for task execution (e.g., in motor imagery cases~\cite{di2018}), mental workload~\cite{ghimatgar2019} , eye movements, attention, decision-making in specific tasks~\cite{craik2019}, and stress, the exploration of perceptual brain decoding with EEG remains relatively limited. Nonetheless, there have been a few recent studies focusing on Perceptual Brain Decoding (PBD) that utilize deep learning methods applied to EEG signals for identifying or reconstructing input stimuli or associated mental imagery. While these developments are promising, given the novelty of this field, there is substantial room for improvement in method performance, the diversity of input considerations, the generalizability of methods on larger datasets, and their analysis. In this research, our primary emphasis lies on reconstructing image stimuli, specifically addressing the task of reconstructing images from EEG data. We collect EEG signals while presenting images from various categories to participants and endeavor to reconstruct images belonging to the same category as the presented stimuli based on the relevant EEG signals.

EEG signals typically exhibit noise, posing a challenge to the task of image synthesis. This challenge can result in the generation of low-quality or irrelevant images if the network fails to effectively learn the conditional mapping. Consequently, a basic generative adversarial network (GAN) architecture may prove insufficient. We aim to address these challenges by enhancing both the quality of synthesized images and their class-specific relevance, an objective that has seen limited exploration in existing literature.

In this study, we employ a combination of a pre-trained convolutional neural network (CNN) and a conditional GAN. The CNN, trained on class-specific EEG signals, serves to extract EEG encodings, which are subsequently used as input for the GAN. Training of the GAN incorporates adversarial loss as well as perceptual loss derived from an intermediate layer of an additional pre-trained image classifier CNN. To assess performance, we utilize metrics such as inception score and class diversity score. Our approach demonstrates superior performance in generating images compared to state-of-the-art methods, as evidenced by improvements in both metrics. 
Our core contributions for this work are summarized as:
\begin{itemize}
    \item We propose to use a conditional GAN based architecture for EEG to image synthesis problem. The class-specific EEG encoding obtained from a pre-trained EEG encoder are used as inputs to GAN. Importantly, since the EEG signals are themselves noisy, we do not provide an additional external noise to the GAN, as is done in existing approaches, and traditionally in GAN.
    
    \item In addition to the standard adversarial loss, a perceptual loss is also used to train the GAN. 
    
    \item We demonstrate that the above contributions lead to an improvement with respect to the class-specific relevance, as measured by the class diversity score. 
\end{itemize}
\section{Related Work}
Traditionally, the EEG signal was mainly used for applications like seizure detection~\cite{chen2014} and that is the reason of the availability of great amount of literature about EEG in medical applications. Recently, the scientists also started working on different applications like perceptual brain decoding, mental load classification etc. Even though a large fraction of works based on EEG classification using deep learning mainly focus on tasks like seizure detection~\cite{chen2014}, recent research also includes applications like event-related potential detection~\cite{parekh2017}, emotion recognition~\cite{chen2019}, mental workload~\cite{di2018}, motor imagery~\cite{he2018} and sleep scoring~\cite{ghimatgar2019} etc., the significant amount of literature is also available on EEG analysis using DL/ML approaches for perceptual brain decoding.

The authors in~\cite{craik2019} briefly discuss the deep learning based important implementations and the results for the EEG classification in the applications like emotion recognition, motor imagery etc. Another attempt to the classification of emotions using EEG signals was successfully done in~\cite{chen2019}. Here, the authors proposed CNN based deep learning framework that works on the combination of time domain and frequency domain features on the DEAP dataset. One more attempt for emotion classification using EEG signals and deep learning was proposed by the authors in~\cite{chen2019 , gao2015}. In these works, the authors proposed the use of machine/deep learning techniques like k-NN, ANN and CNN for the classification of emotions with EEG signals.  The authors in~\cite{bashivan2015learning} worked with the combination of different deep learning architectures like CNN and LSTM for the task of EEG classification for motor-imagery tasks. One interesting applications of image annotations using EEG signals and deep learning was suggested in the work~\cite{parekh2017}. 

In continuation of perceptual decoding, several attempts have been made in order to classify EEG signals corresponding to different visual stimulus. One of a very recent approach that deals with the EEG classification for the task of visual perception is given by~\cite{Tiruppatur}. In this work, the authors proposed a deep learning network for the classification of EEG signals while the signals have been captured by Emotiv Epoc (14- channels) device. Parallel to this work, the authors of~\cite{jolly2019universal} also proposed a GRU based deep learning approach to classify the EEG signals from the ThoughtViz dataset~\cite{Tiruppatur}. The authors in~\cite{mishra2021eeg} proposed siamese network based technique for EEG classification on Thoughtviz dataset. One more attempt for the classification of EEG signals corresponding to MNIST digits was also reported in the work by\cite{mishra2021visual} This work is motivated from~\cite{Tiruppatur} where the authors used GAN based techniques for generating visual stimulus back from the EEG signals. 

\section{Dataset Details}
\begin{figure}[ht]
    \centering
    \includegraphics[scale = 0.38]{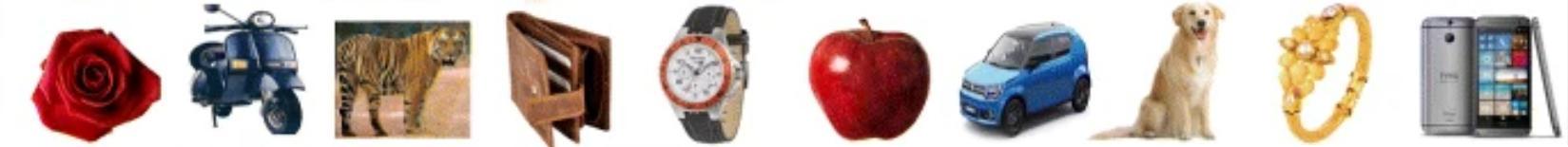}
    \caption{Sample images of visual stimuli}
    \label{sample_img}
\end{figure}

The dataset for this work acquired from Kumar et al.’s work~\cite{kumar2018envisioned}. This dataset contains EEG recordings from 23 volunteers who were shown 10 examples of visual stimuli from 10 different object classes from the ImageNet dataset. Few sample images of these stimuli are shown in Fig~\ref{sample_img}. Each EEG signal is recorded for the duration of 10 seconds. 

Tirupattur et al., 2018~\cite{Tiruppatur} released this dataset after dividing the EEG recordings into smaller parts with a window size of 32 samples and an overlap of 8 samples. This EEG data is collected using Emotiv EPOC headset. This dataset doesn't contain an exact one-to-one mapping of EEG parts to particular images. However, the mapping between the EEG parts and the class of objects is available. 
Hence, the task involves synthesizing some image of an object corresponding to the input EEG signals acquired when that class of images were shown~\cite{Tiruppatur}. 

\begin{figure}[ht]
    \centering
    \includegraphics[scale = 0.6]{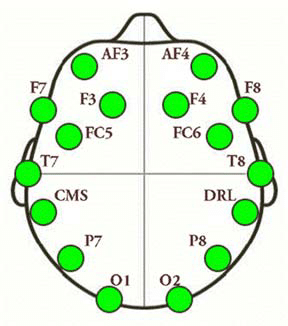}
    \caption{Different brain locations for Emotiv EPOC device~\cite{epoc_diag}}
    \label{epoc}
\end{figure}

The different brain locations for emotiv epoc is given in Fig.~\ref{epoc}. This EEG capturing device contains 14 channels with the sampling frequency 128 Hz. As a first step, the pre-processing of this raw EEG data is done using a sliding window of 32 samples with overlapping of 8 samples. 

\section{Methodology}
\begin{figure}[h!]
    \centering
    \includegraphics[scale = 0.4]{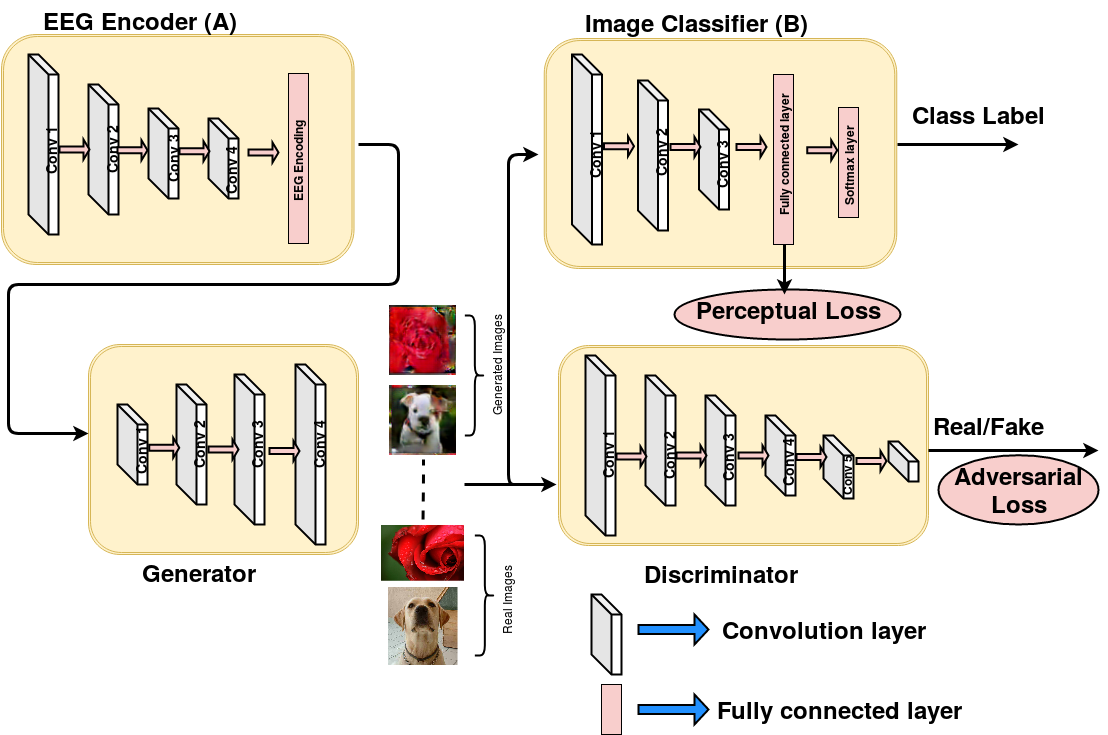}
    \caption{Block diagram of the proposed methodology }
    \label{fig:my1}
\end{figure}

To generate class-specific images  corresponding to the EEG signals, we use generative adversarial network~\cite{goodfellow2014generative} (GAN) as the backbone of our proposed architecture.   Fig.~\ref{fig:my1} shows the network architecture where the two additional blocks - A) EEG encoder and, B) image classifier - are used along with a traditional GAN. The description of each block is given in the subsequent subsections.

\subsection {GAN}
 GAN is a well-studied deep learning based network which is made up of two blocks - Generator ($G$) and Discriminator ($D$), both are trained in an adversarial fashion.  The generator  generates images from the random noise input ($z$) and the discriminator tries to classify the generated image as fake or real. The adversarial learning pushes the generator to generate images that are very close to real images ($x$)  by fooling the  discriminator. The generator learns to produce the  samples close to the real image distribution $(p_{data})$ by mapping the input noise ($z$) from a known distribution $(p_{z})$ to the real image distribution.
The adversarial loss function of the basic GAN network  is
\begin{equation}
L1: ~~~~~\underset{D}{max} \hspace{2pt}\underset{G}{min} V_{D}(D, G) = {\underset{D}{max} \hspace{2pt}\underset{G}{min}} \big(\mathbb{E}_{x\sim p_{data}}[log(D(x))] + \mathbb{E}_{z\sim p_{z}}[log(1 - D(G(z)))] \big).
\end{equation}

\noindent In this  work, our task is  to generate class-specific visual stimulus, i.e., the class labels of the EEG signals and the generated images should be the same. A basic GAN architecture is not sufficient for this task. Due to the inability of basic GAN network to produce class-specific images we focus on the conditional GAN paradigm where the condition is fed to the generator through input signals. Here, we give this condition as the class-specific EEG encodings (from EEG encoder, details are in section-4.2). Motivated from the AC-GAN~\cite{odena2017conditional} architecture, we  use an additional image classifier for class-specific image generation. In this work, we observe that the EEG signals are themselves noisy in nature~\cite{pijn1991chaos}. Hence, their encodings (details in the subsection 4.2), can be seen as a combination of signal and noise. Thus, we use these encodings solely as the input of the generator, and do not use an additional noise input, as is usually done for GANs. We believe that the additional noise might deteriorate the performance of the generator for generating class-specific images. Indeed, we quantitatively demonstrate that our approach yields more relevant images as compared to the state-of-the-art which includes the additional noise factor.

\subsection{C-Former EEG Encoder}
The EEG signals are typically noisy, leading to the difficulties in training a GAN network if fed directly in  their raw form.  In addition, these signals are also high dimensional  which can lead to suppression of class-specific  discriminative information.  Hence, using raw EEG signals as input to GAN is not a good idea. Inspired by Tiruppatur et. al~\cite{Tiruppatur}, we transform raw EEG signals to low-dimensional encoding vectors using a Transformer-encoder (Block A in Fig. \ref{fig:my1}) based EEG encoder which is trained to generate class-specific embeddings. 
Transformer architecture essentially comprises an encoder and a decoder module. However, if the required application involves finding features for a specific task, one can work solely with the encoder module. In this work, since our objective is to derive EEG features for EEG classification, we exclusively utilize the encoder module.

Having said that, in constructing the EEG encoder, we employ a network architecture known as the C-former, which integrates elements of convolution neural networks (CNNs), self-attention modules, and a fully-connected classifier. This network architecture is motivated from a recent research work~\cite{song2022eeg}. Below we discuss the relevant details of all the different modules. 
\begin{itemize}
	\item \textbf{Convolution module:} Inspired by the methodologies discussed in~\cite{Tiruppatur} and~\cite{mishra2021visual}, we design the convolution module by decomposing the two-dimensional convolution operator into separate one-dimensional temporal and spatial convolution layers. In the initial layer, k kernels of dimensions (1, 5) are utilized with a stride of (1, 1), indicating that the convolution operation is applied across the time dimension. Subsequently, the layer maintains k kernels of size (ch, 1) with a stride of (1, 1), where \textbf{ch} denotes the number of electrode channels in EEG data. This layer acts as a spatial filter, capturing the interactions among different electrode channels. To improve the training process and address overfitting concerns, we integrate batch normalization. 
	
	\item \textbf{Self-attention module:}  In this specific part of our work, we introduce self-attention to understand the broader temporal relationships among EEG features. This helps compensate for the limited coverage of the convolution module. We take the tokens organized in the previous step, give them new shapes called query (Q), key (K), and value (V) through a linear transformation. We then measure the relationship between these tokens using a dot product between Q and K~\cite{song2022eeg}.  We also utilize a multi-head strategy to enhance the diversity of representations. The tokens are evenly split into h (h = 8 for this work) segments, each fed into the self-attention module independently. 
	The outputs from these segments are then combined to form the final output of the module. Output of this module is fed as an input to the feed-forward input layer.

	\item \textbf{Classifier module:} Finally, we incorporate two fully-connected layers to serve as the classifier module. This module produces an $M$-dimensional vector (here, $M$ is the number of classes) following the application of the softmax function. As a loss function, we use cross-entropy.
\end{itemize}

\begin{figure}[h!]
	\centering
	\includegraphics[scale = 0.45]{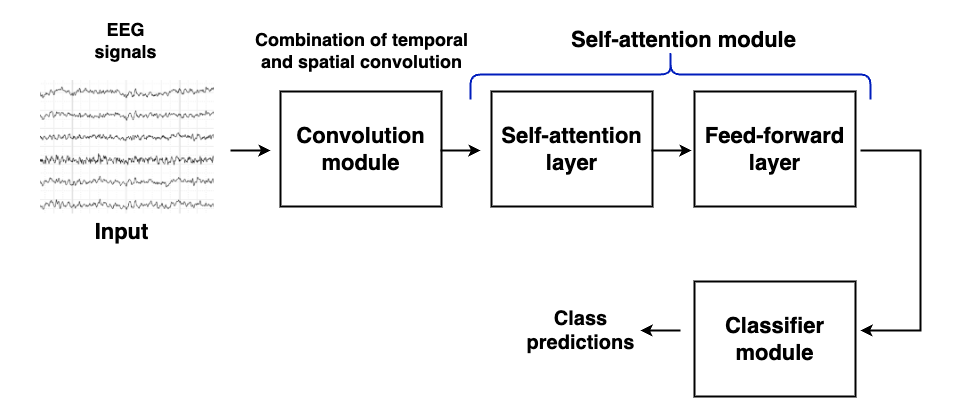}
	\caption{C-former EEG classifier }
	\label{c_form}
\end{figure}
The complete architecture is given in Figure-~\ref{c_form}. For getting the embeddings from this C-Former network, 
\begin{itemize}
	\item Firstly, we pre-trained it on the ThoughtViz~\cite{Tiruppatur} EEG dataset for classification.
	\item After training the C-former network, we removed the last softmax layer to obtain embeddings from the second-last fully connected layer. 
	
\end{itemize}

\subsection{Image Classifier}
We  also use an additional  CNN based image classifier (B in Fig. \ref{fig:my1}) which computes the classification loss on the generated images. The classifier (C) is trained separately from scratch on real images from ImageNet~\cite{deng2009imagenet} dataset with their corresponding class labels. It consists of  3 convolutional layers with  $3\times3$ kernel size, with each layer having 32, 32 and 64 filters, respectively. The convolutional layers are followed by two fully connected layers with 128 and 64 neurons, respectively. We use softmax activation function at the output layer of the network. The network is trained only for the 10 ImageNet classes whose images were shown at the time of EEG signal acquisition. We use 80\% images per class for training the network and remaining 20\% for  testing and achieved an overall test accuracy of 81\%.

\noindent For producing the class-specific images from GAN, we make use of the classification loss of image classifier in addition to the GAN  adversarial loss. The classifier loss is
\begin{equation}
L2:~~~~~~~ \mathbb{E}_{z \sim p_z} log(C(G(z))),
\end{equation}
where z denotes the EEG encoding to the generator input.
The introduction of this additional image classifier is motivated from AC-GAN~\cite{odena2017conditional} architecture, where the classification loss forces the GAN network to produce more class-specific images. Along with this, we also propose to use a perceptual loss for generating more realistic images from the generator. The perceptual loss between real image and generated image is given as 
\begin{equation}
 L3:~~~~~~~  \frac{1}{N} \sum_{i=1}^{N} ~ |C_l(x) - C_l(G(z))|.
\end{equation}
Here, $C_l(x)$ and $C_l(G(z))$ are the respective embeddings of real image and generated image from the $l^{th}$ layer of the image classification network. The perceptual loss is essentially a feature-level loss computed at a sufficiently deep features extracted from a network. It helps the network to minimize the perceptual quality difference between the real images and generated images.


\subsection{Loss Function \&  Network Training}
The generator part of the network converts EEG encodings into the RGB images. The generator architecture consists of a combination of convolution layers  and upsampling layers. The generator takes the input of size 100 $\times$ 1 and generates the output of size  64 $\times$ 64 $\times$ 3. The discriminator is a simple image classifier type of network consists of 5 convolution layers. It takes a $64 \times 64 \times 3$ image   as an input and classifying it as real/fake. These real images are from the ImageNet dataset. Traditionally, the loss functions to train the GAN network is the standard adversarial loss function. But here we make use of classifier and perceptual loss along with the adversarial  loss to minimize the overall cost function. 
\begin{equation}
L_{total} =  L1 + L2 + L3.
\end{equation}

We start the training of this complete GAN network with a batch size of 100 and a learning rate of $10^{-5}$. The generated images and the real images are fed to the classifier network for the calculating the classification loss and perceptual loss. The overall flow of this work is given in Fig. \ref{fig:my1}.

\section{Experiments \& Results}

\subsection{Evaluation Measure}
To evaluate the performance of proposed approach we use two different performance metric scores: inception score~\cite{salimans2016improved} and class diversity score~\cite{ben2018multi} . 
\begin{itemize}
    \item \textbf{Inception Score:} It is a metric to evaluate the quality of generated images from the generator. The inception score can be calculated using the given formula:
    \begin{equation*}
        IS(G) = exp\big(\mathbb{E}_{G(z) \sim p_{g}} D_{KL}( p(y|G(z))~||~ p(y))\big)
    \end{equation*}
    
    Here, $G(z)$ represents a generated sample, sampled from a distribution ($p_{g}$).
    \\ $D_{KL}(p||q)$ is the KL-Divergence between the distributions $p$ and $q$. 
    \\ $p(y|G(z))$ is the conditional class distribution.
    \\ $p(y)$ = $\int_{G(z)}^{} p(y|G(z)) p_{g}(G(z))$ is the marginal class distribution. 
    \\ For high quality generated images, the inception score should be high. 
    
    \item \textbf{Class Diversity Score:} To compute the class-diversity score of generated images, we use a pre-trained image classifier. The obtained predictions from the classifier for the generated images are in the form of one hot encoding vectors (C(G(z))). Class diversity of each class is computed by taking the entropy of the average of the one-hot prediction vectors. The class diversity score is defined as
    \begin{equation*}
        Score = \frac{1}{log(N)}H\bigg{(}\frac{1}{\left | X \right |} \sum_{G(z) \in X} C(G(z))\bigg{)}.
    \end{equation*}
    Here, $N$ denotes the total number of classes, $|X|$ represents total number of generated samples from EEG signals of a particular class, and $C$ is the classifier. The range of diversity score is in between 0 to 1. A good diversity score should be as low as possible. A high diversity score indicates higher irrelevance, i.e., that many images of other classes are also being synthesized than that of the input EEG signals.
\end{itemize}

\begin{table}[h!]
\centering
\caption{Comparison of Diversity score with ThoughtViz~\cite{Tiruppatur}}
\begin{tabular}{|c|c|c|c|}
\hline
\textbf{S.No} & \textbf{\begin{tabular}[c]{@{}c@{}}Object \\ Class\end{tabular}} & \textbf{\begin{tabular}[c]{@{}c@{}}Diversity Score\\ C-former encoder\end{tabular}} & \textbf{\begin{tabular}[c]{@{}c@{}}Diversity Score on\\ ThoughtViz\end{tabular}} \\ \hline
1             & Apple                                                            & 0.7213                                                                              & 0.8497                                                                           \\ \hline
2             & Car                                                              & 0.5621                                                                              & 0.8391                                                                           \\ \hline
3             & Dog                                                              & 0.6327                                                                              & 0.6965                                                                           \\ \hline
4             & Gold                                                             & 0.5391                                                                              & 0.6561                                                                           \\ \hline
5             & Mobile                                                           & 0.7252                                                                              & 0.8541                                                                           \\ \hline
6             & Rose                                                             & 0.5843                                                                              & 0.8309                                                                           \\ \hline
7             & Scooter                                                          & 0.5915                                                                              & 0.6309                                                                           \\ \hline
8             & Tiger                                                            & 0.7594                                                                              & 0.9068                                                                           \\ \hline
9             & Wallet                                                           & 0.7553                                                                              & 0.8153                                                                           \\ \hline
10            & Watch                                                            & 0.6297                                                                              & 0.8215                                                                           \\ \hline
11            & Mean                                                             & 0.6501                                                                              & 0.7897                                                                           \\ \hline
\end{tabular}
\label{tab1}
\end{table}
\subsection{Results}
The comparison of diversity score on the test data with the existing state-of-the-art work~\cite{Tiruppatur} for different classes is given in Table~\ref{tab1}. To compute the class diversity score from~\cite{Tiruppatur}, we use the released generator by the authors from the link (https://github.com/ptirupat/ThoughtViz). It is evident from Table~\ref{tab1} that we achieve a low diversity score of all the visual classes as compared to that in~\cite{Tiruppatur}. This means that our approach is producing significantly class specific images. 

We also compared our results with the inception score of recent suggested approaches. Typically, the inception score is calculated w.r.t a large number of images. Since, in this case we have only 5706 images corresponding to test signals. Hence for statistical reliability, we consider two different conditions for calculating inception score.
\begin{itemize}
    \item Condition 1: Inception score on combined generated images w.r.t train and test EEG signals (total 50000).
    \item Condition 2: Inception score on generated images w.r.t 5706 test EEG signals only. 
\end{itemize}
All the comparisons are listed in Table 2. From results it is clear that the inception score of our work is almost similar to the state of the art method. 

\section{Conclusion}
In this study, we introduce a method aimed at generating images from EEG signals collected during a perceptual brain decoding task. We hypothesize that the inclusion of additional noise in the EEG encodings may hinder the performance of the generator in synthesizing images specific to particular classes. As a solution, we propose the incorporation of perceptual loss to enhance the generation of realistic images. 

\begin{table}[h!]
\centering
\caption{Comparison of Inception Score from different methods}

\begin{tabular}{cccll}
\cline{1-3}
\multicolumn{1}{|c|}{\textbf{S. No}} & \multicolumn{1}{c|}{\textbf{Method}}                                                                  & \multicolumn{1}{c|}{\textbf{Inception Score}} &  &  \\ \cline{1-3}
\multicolumn{1}{|c|}{1}              & \multicolumn{1}{c|}{AC-GAN~\cite{odena2017conditional}   }                                                                       & \multicolumn{1}{c|}{4.9}                      &  &  \\ \cline{1-3}
\multicolumn{1}{|c|}{2}              & \multicolumn{1}{c|}{\begin{tabular}[c]{@{}c@{}}AC-GAN~\cite{odena2017conditional} \\ (one-hot encoding)\end{tabular}}             & \multicolumn{1}{c|}{3.10}                     &  &  \\ \cline{1-3}
\multicolumn{1}{|c|}{3}              & \multicolumn{1}{c|}{\begin{tabular}[c]{@{}c@{}}ThoughtViz~\cite{Tiruppatur}\\ (on 50000 generated images : similar to condition-1)\end{tabular}} & \multicolumn{1}{c|}{5.43}                     &  &  \\ \cline{1-3}
\multicolumn{1}{|c|}{4}              & \multicolumn{1}{c|}{\begin{tabular}[c]{@{}c@{}}ThoughtViz~\cite{Tiruppatur} \\ (only on 5706 test images : similar to condition-2)\end{tabular}} & \multicolumn{1}{c|}{4.82}                     &  &  \\ \cline{1-3}
\multicolumn{1}{|c|}{5}              & \multicolumn{1}{c|}{\textbf{\begin{tabular}[c]{@{}c@{}}Ours \\ Condition:1\end{tabular}}}             & \multicolumn{1}{c|}{\textbf{5.1}}             &  &  \\ \cline{1-3}
\multicolumn{1}{|c|}{6}              & \multicolumn{1}{c|}{\textbf{\begin{tabular}[c]{@{}c@{}}Ours\\ Condition:2\end{tabular}}}              & \multicolumn{1}{c|}{\textbf{4.62}}             &  &  \\ \cline{1-3}
\multicolumn{1}{l}{}                 & \multicolumn{1}{l}{}                                                                                  & \multicolumn{1}{l}{}                          &  & 
\end{tabular}
\end{table}

\section{Acknowledgment}
This work has been partially supported by SERB, Government of India, under Project CRG/2022/007117. 
  

\bibliography{report} 
\bibliographystyle{spiebib} 

\end{document}